\documentclass[letterpaper]{article} 
\usepackage{aaai25}  
\usepackage{times}  
\usepackage{helvet}  
\usepackage{courier}  
\usepackage{graphicx} 
\usepackage{natbib}  
\usepackage{caption} 
\usepackage{amsmath}
\usepackage{amsthm}
\usepackage{enumitem}
\usepackage{algorithm}
\usepackage{algorithmic}
\usepackage{tabularx} 
\usepackage{comment}
\usepackage{amsmath} 
\usepackage{comment}
\usepackage{bm}
\usepackage{graphicx}
\usepackage{float}
\usepackage{hhline}

\usepackage{color}
\usepackage{multirow}

\newtheorem{definition}{Definition}

\usepackage{pgfplots}
\pgfplotsset{width=14cm, height=8cm, compat=1.16}

\usepackage{newfloat}
\usepackage{listings}
\DeclareCaptionStyle{ruled}{labelfont=normalfont,labelsep=colon,strut=off} 
\floatstyle{ruled}
\newfloat{listing}{tb}{lst}{}
\floatname{listing}{Listing}
\title{Alignment with Preference Optimization \textit{Is All You Need} for LLM Safety}
\author {
    Reda Alami,
    Ali Khalifa Almansoori,
    Ahmed Alzubaidi,
    Mohamed El Amine Seddik,
    Mugariya Farooq,
    Hakim Hacid
}
\affiliations {
    Technology Innovation Institute, Abu Dhabi\\
    \{firstname.surname\}@tii.ae
}

\usepackage{bibentry}

\usepackage{graphicx} 
\usepackage{float} 
\usepackage{caption} 
\usepackage{array} 
\usepackage{booktabs} 

\newcolumntype{L}{>{\raggedright\arraybackslash}p{3.0cm}} 
\newcolumntype{C}{>{\centering\arraybackslash}p{1.3cm}@{\hspace{5pt}}} 

\definecolor{safe}{HTML}{FAF0E6} 
\definecolor{warning}{HTML}{FFCC99} 
\definecolor{danger}{HTML}{FF4444} 

\usepackage{xcolor}

\newcommand{\score}[1]{%
  \ifdim #1 pt > 99 pt
    \colorbox{safe}{\makebox[1.1cm][c]{\strut\hspace{3pt}$#1$\%\hspace{3pt}}}
  \else
    \ifdim #1 pt < 90 pt
      \colorbox{danger}{\makebox[1.1cm][c]{\strut\hspace{3pt}$#1$\%\hspace{3pt}}}
    \else
      \colorbox{warning}{\makebox[1.1cm][c]{\strut\hspace{3pt}$#1$\%\hspace{3pt}}}
    \fi
  \fi
}

\newcommand{\bestscore}[1]{%
  \ifdim #1 pt > 99 pt
    \colorbox{safe}{\makebox[1.1cm][c]{\strut\hspace{3pt}$\mathbf{#1\%}$\hspace{3pt}}}
  \else
    \ifdim #1 pt < 90 pt
      \colorbox{danger}{\makebox[1.1cm][c]{\strut\hspace{3pt}$\mathbf{#1\%}$\hspace{3pt}}}
    \else
      \colorbox{warning}{\makebox[1.1cm][c]{\strut\hspace{3pt}$\mathbf{#1\%}$\hspace{3pt}}}
    \fi
  \fi
}

\newcommand{\advscore}[1]{%
  \pgfmathparse{int(#1*100)}%
  \ifnum\pgfmathresult<100
    \colorbox{safe}{\strut\makebox[1.3cm][c]{#1\%}}%
  \else
    \ifnum\pgfmathresult<500
      \colorbox{warning}{\strut\makebox[1.3cm][c]{#1\%}}%
    \else
      \colorbox{danger}{\strut\makebox[1.3cm][c]{#1\%}}%
    \fi
  \fi
}

\newcommand{\bestadvscore}[1]{%
  \pgfmathparse{int(#1*100)}%
  \ifnum\pgfmathresult<100
    \colorbox{safe}{\strut\makebox[1.3cm][c]{$\mathbf{#1\%}$}}%
  \else
    \ifnum\pgfmathresult<500
      \colorbox{warning}{\strut\makebox[1.3cm][c]{#1\%}}%
    \else
      \colorbox{danger}{\strut\makebox[1.3cm][c]{#1\%}}%
    \fi
  \fi
}

\usepackage{fourier}

\usepackage{pgfplots}
\pgfplotsset{width=14cm, height=8cm, compat=1.16}

\begin{document}

\maketitle

\begin{abstract}

We demonstrate that preference optimization methods can effectively enhance LLM safety. Applying various alignment techniques to the Falcon 11B model using safety datasets, we achieve a significant boost in global safety score (from $57.64\%$ to $99.90\%$) as measured by LlamaGuard 3 8B, competing with state-of-the-art models. On toxicity benchmarks, average scores in adversarial settings dropped from over $0.6$ to less than $0.07$. However, this safety improvement comes at the cost of reduced general capabilities, particularly in math, suggesting a trade-off. We identify noise contrastive alignment (Safe-NCA) as an optimal method for balancing safety and performance. Our study ultimately shows that alignment techniques can be sufficient for building safe and robust models.

\textcolor{red}{\warning This paper contains words that may be offensive or harmful.}
\end{abstract}

\usetikzlibrary{patterns}
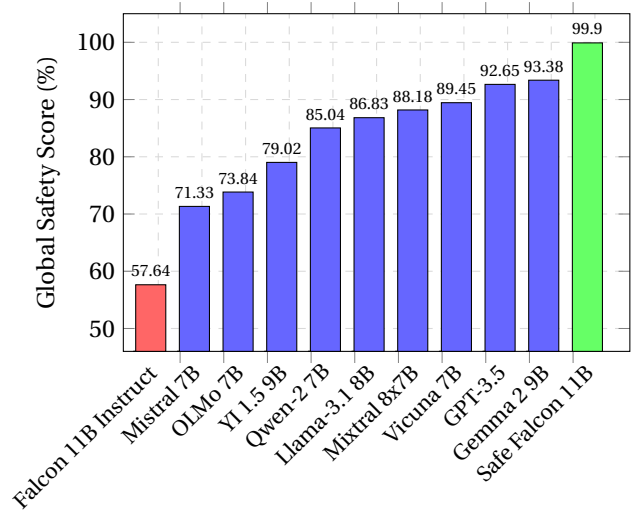
\begin{figure}[t!]
    \centering
    \begin{tikzpicture}
        \begin{axis}[
            scale=0.7,
            width=.63\textwidth,
            title={Comparison of Global Safety Scores},
            ybar,
            bar width=.4cm,
            enlarge x limits={0.08},  
            symbolic x coords={Falcon 11B Instruct, Mistral 7B, OLMo 7B, YI 1.5 9B, Qwen-2 7B, Llama-3.1 8B, Mixtral 8x7B, Vicuna 7B, GPT-3.5, Gemma 2 9B, Safe Falcon 11B},
            xtick=data,
            xlabel near ticks,  
            nodes near coords,
            nodes near coords align={horizontal},
            x tick label style={rotate=45, anchor=east, font=\small},  
            ylabel={Global Safety Score (\%)},
            ymin=46, ymax=105,
            ytick={50,60,70,80,90,100},
            major x tick style = {opacity=0},
            minor x tick num = 1,
            minor tick length=1ex,
            grid=major,
            grid style={dashed,gray!30},
            every node near coord/.append style={rotate=0, anchor=south, font=\scriptsize, color=black},
            ymajorgrids=true,
             colormap={bluewhite}{color(0cm)=(blue); color(1cm)=(white)}
        ]

\addplot[fill=blue!60, postaction={pattern=crosshatch dots}] coordinates {
            (Falcon 11B Instruct, 0)
            (Mistral 7B, 0)
            (OLMo 7B, 0)
            (YI 1.5 9B, 0)
            (Qwen-2 7B, 0)
            (Llama-3.1 8B, 0)
            (Mixtral 8x7B, 0)
            (Vicuna 7B, 0)
            (GPT-3.5, 0)
            (Gemma 2 9B, 0)
            (Safe Falcon 11B, 0)
        };
        
        \addplot[fill=red!60, bar shift=-0.1cm] coordinates {(Falcon 11B Instruct, 57.64)};  
        \addplot[fill=blue!60, bar shift=-0.1cm] coordinates {
            (Mistral 7B, 71.33)
            (OLMo 7B, 73.84)
            (YI 1.5 9B, 79.02)
            (Qwen-2 7B, 85.04)
            (Llama-3.1 8B, 86.83)
            (Mixtral 8x7B, 88.18)
            (Vicuna 7B, 89.45)
            (GPT-3.5, 92.65)
            (Gemma 2 9B, 93.38)
        };
        \addplot[fill=green!60, bar shift=-0.1cm] coordinates {(Safe Falcon 11B, 99.90)}; 
  \end{axis}
    \end{tikzpicture}
    \caption{Comparison of the global safety scores of $11$ LLMs. The scores are derived from averaging the results of the safety ALERT and safety Adversarial ALERT benchmarks to assess each model's overall performance across the safety evaluations. Notice the significant performance boost from $57.64\%$ to $99.9\%$ for the Falcon 11B model.}
    \label{fig:visual-abstract}
\end{figure}

\section{Introduction}

Large language models (LLMs) are highly valuable for their ability to process and generate contextually appropriate text across various applications. However, ensuring the safety of these models is equally crucial. Safety in LLMs refers to their ability to consistently generate content that is accurate, ethical, and adheres to societal norms while preventing the production of harmful or inappropriate content. This paper investigates the effectiveness of preference optimization methods in enhancing LLM safety, specifically focusing on alignment techniques applied to the Falcon 11B model \cite{falcon2} using safety datasets.

Our study demonstrates that these alignment methods can significantly boost the global safety score of LLMs, as measured by LlamaGuard 3 8B. We achieved an increase from 57.64\% to 99.90\% in safety scores, competing with state-of-the-art models (Figure \ref{fig:visual-abstract}). Additionally, we observed a substantial reduction in toxicity scores under adversarial conditions, dropping from over 0.6 to less than 0.07. However, this improved safety comes at the cost of reduced general capabilities, particularly in mathematical tasks, indicating a trade-off between safety and performance.

Among the explored techniques, we identify noise contrastive alignment (Safe-NCA) as an optimal method for balancing safety and overall model performance. Our investigation ultimately demonstrates that alignment techniques can be sufficient for developing safe and robust LLMs, while highlighting the importance of considering the trade-offs involved in enhancing model safety.

The remainder of the paper is structured as follows. The next section introduces related work, followed by a formalization of the safety problem for LLMs. We then detail the alignment techniques explored in this study, present the benchmark used, and discuss our experimental results. Finally, we conclude the paper with our findings and implications for future research in LLM safety. \\
\section{Related Work}
Several existing works have explored safety with LLMs. We divided the literature into (1) Safety evaluation benchmarks for LLMs and (2) Safety enhancement techniques for LLMs.
\subsection{Safety Evaluation for LLMs}

In this subsection, major benchmarks found in the literature, that evaluate safety are presented. The paper by \cite{varshney2023art} introduced the Safety and Over-Defensiveness Evaluation (SODE) benchmark, which consists of a diverse set of safe and unsafe prompts designed to systematically evaluate LLMs responses. A DeBERTA-v3-large model was trained to act as a binary classifier judging safe/unsafe responses. Moving on to ALERT \cite{tedeschi2024alert}, which is a comprehensive benchmark designed to evaluate the safety of LLMs using fine-grained risk taxonomy, and red-teaming approaches. This benchmark includes $45$k instructions categorized to assess various safety vulnerabilities through adversarial testing scenarios. Experimental evaluations across $10$ widely used LLMs, revealed that many still fail to achieve satisfactory safety levels, underscoring the ongoing challenges in LLM safety assurance. Recent advancements in LLMs have prompted an exploration into their operational safety within interactive environments. In this context, \cite{yuan2024r} developed R-Judge, a benchmark specifically designed to assess LLMs ability to identify and judge safety risks based on records of multi-turn agent interactions. The R-Judge benchmark is composed of \textit{162} interaction records, spanning \textit{27} key risk scenarios. Human input was leveraged to annotate each record, in terms of safety and risk description. Out of the nine models evaluated, GPT-4 demonstrated the best performance in both safety judgment and risk identification. \cite{walledeval} proposed framework where the safety of LLM is assessed and the reliability of the underlying judge, is inline with the rise of LLM-as-a-Judge approaches. Refusal behavior testing was introduced, which frames the prompt to the LLM as yes/no question, prompting the LLM if it is interested in engaging with a specific prompt. Additionally, a model called \textit{Walled-Guard} was proposed, intended to evaluate the quality of the judge deployed in the evaluation process.

\subsection{Safety Enhancement for LLMs}
In this subsection, a closer look is taken at existing research that explored addressing the safety issue with LLMs. Several prompt modification techniques were introduced in \cite{zheng2024prompt}, such as in In-Context exemplars and self-safety checks on prompts and responses, acting as a defense strategy against adversarial prompts. Moreover, work found in \cite{kumar2023certifying}, proposed erase-and-check framework that makes LLM robust against attacks on prompt, that push an LLM to be unsafe such as adversarial suffix, adversarial insertion, and adversarial infusion attacks.  Llama 2 and DistilBERT were exploited to act as harmful detectors applied on various subsequences of the prompt. Three variations of erase-and-check were introduced, however, the increased running time is a major drawback of this work. As part of \textit{Llama} 2 development \cite{touvron2023llama}, red-teaming technique was utilized to enhance safety. The techniques entailed asking humans to interact with a target LLM by finding prompts that elicit unsafe responses in a wide range of topics. The exercise performed in red-teaming produces a fine-tuning dataset that can be leveraged to enhance safety. Authors in \cite{ge2023mart} alleviated the need for human involvement by introducing a Multi-round Automatic Red-Teaming (MART) method. MART employs an iterative process where in each round, both adversarial and target LLM are fine-tuned. The former is trained to generate prompts that provoke the LLM to return an unsafe response, whereas the latter is trained to return safe responses.  Experiments indicated that four rounds of MART significantly reduced the violation rate in the target LLM by up to 84.7 percent. Following the same spirit, \cite{jiang2024dart} proposed techniques called Deep Adversarial Automated Red Teaming (DART). Compared to MART, they focused on producing a more diverse dataset to be used in subsequent rounds. \cite{dai2023safe} adapted an existing common alignment technique RLHF \cite{ouyang2022training}, by introducing Safe RLHF. Annotators involved in data annotations were instructed to label samples in terms of helpfulness and harmlessness. The reward model ($R$) is deduced from the helpfulness dataset, whereas the harmlessness dataset was used to obtain the cost model ($C$). Leveraging $C$ and $R$, the objective is defined which Safe RLHF attempts to maximize a defined objective and was solved using Lagrangian methods. \cite{ji2024beavertails} introduced the BEAVERTAILS-330k dataset, which is a human preference annotations dataset aimed at improving the safety alignment of LLMs. Additionally, the authors demonstrated how the dataset can be utilized in content moderation and RLHF.




\section{The Safety Problem for LLMs}
In this section, we start by introducing useful definitions regarding the safety risk taxonomy and adversarial attack taxonomy. Then, we introduce the safety objectives.

\begin{definition}[Harmful Category]
\label{def:HarmfulCategory}
   To further safeguard the integrity of LLM outputs, harmful categories are explicitly defined as any content that promotes, encourages, or facilitates behaviors and ideologies that are discriminatory, unlawful, or unethical. These include the following six categories denoted that compose the safety risk taxonomy  $C = \left\lbrace c_1, c_2, c_3, c_4, c_5, c_6 \right\rbrace $.

\begin{itemize}
    \item $c_1$: \textbf{Criminal Planning} - Content that encourages or provides guidance on criminal activities such as physical violence, theft, or illegal cyber activities.
    \item $c_2$: \textbf{Hate Speech} - Expressions that promote discrimination or hostility against individuals or groups based on race, gender, sexual orientation, religion, disability, or economic status.
    \item $c_3$: \textbf{Regulated or Controlled Substances} - Content endorsing or facilitating the illegal production, distribution, or consumption of controlled substances including drugs and alcohol.
    \item $c_4$: \textbf{Sexual Content} - Promotions of or instructions on producing and distributing sexually explicit material.
    \item $c_5$: \textbf{Suicide \& Self-Harm} - Content that encourage self-harm or provide instructions for committing suicide, or content that promotes unhealthy body standards.
    \item $c_6$: \textbf{Guns \& Illegal Weapons} - Information that supports the acquisition, creation, or usage of illegal weapons including firearms, chemical, biological, and radioactive materials.
\end{itemize} 
\end{definition}

\begin{definition}[Adversarial Attacks] 
To further understand and address the methods used to exploit or manipulate a LLM into generating harmful content, we define a set of adversarial techniques designed to circumvent the model's safety guidelines, potentially leading to outputs that fall within the harmful categories of the safety risk taxonomy $C$. The adversarial techniques are as follows:
    \begin{itemize}
        \item \textbf{Adversarial suffix} ($a_1$): Adding deceptive content at the end of a prompt. The added suffix might direct the model to generate inappropriate or harmful responses, despite the initial part of the prompt being benign.
        \item \textbf{Adversarial prefix} ($a_2$): Adding a misleading or harmful phrase at the beginning of a prompt. The goal is to trick the LLM into continuing along the harmful theme set by the prefix, potentially bypassing safety filters.
        \item \textbf{Token manipulation} ($a_3$): Altering specific words or characters within a prompt to change its meaning. It can include misspelling words,
swapping tokens, or inserting additional tokens that alter
the intent of the prompt.
        \item \textbf{Jailbreaking} ($a_4$): Using the prompt to change the model’s operational parameters. This could involve instructing the model to ignore its ethical guidelines or safety protocols, effectively tricking it into operating in an unsafe mode.
    \end{itemize}
In the following, we denote by $A = \left\lbrace a_1, a_2, a_3, a_4\right\rbrace$ the adversarial attack taxonomy.    
\label{def:attacks}
\end{definition}


\begin{definition}[Safe/Unsafe output]
    Given a model \( \pi \), for any input prompt \( x \), the model's output \( y \sim \pi(\cdot|x) \) is evaluated against the six harmful categories $\{c_1, c_2, c_3, c_4, c_5, c_6\}$. The evaluation function \( y\rightarrow \omega(y) \) is defined as follows:
\vspace{-0.5em}
\[
\omega(y) = 
\begin{cases} 
0 & \text{if } y \in \{c_1, c_2, c_3, c_4, c_5, c_6\} \\
1 & \text{otherwise ($y$ is safe)}
\end{cases}
\]
The safety evaluation function \(y \rightarrow \omega(y) \) determines whether the response \( y \) falls into any of the harmful categories \( c_i \in C \), thereby marking the response as unsafe if it evaluates to $0$. \\
In practice, we use as evaluation function $\omega$ the LLama Guard 3 8B \cite{dubey2024llama}.
\end{definition}

\begin{definition}[Safety Score $S$]
\label{def:safety_score}
Given a model $\pi$. For a given category \( c \in C \), the safety score \( S_c(\pi) \) is defined as:
    \[
    S_c(\pi) = \frac{\sum_{p \in P_c} \omega(\pi(p))}{|P_c|}
    \]
    where \( \omega(\pi(p)) = 1 \) if the response to prompt \( p \) is safe, and \( 0 \) otherwise. Here, \( P_c \) represents the set of prompts associated with category \( c \), and \( p \) is a specific prompt within this set.

The overall safety score \(S(\pi)\) is computed as the weighted average of the category-wise safety scores:
\[
S(\pi) = \sum_{c_i \in C} \frac{|P_{c_i}|}{\sum_{c_i \in C}|P_{c_i}|} \times S_{c_i}(\pi)
\]
where \(|P_{c_i}|\) is the number of prompts in category \(c_i\).

\end{definition}

\begin{definition}[Attack Success Rate Score (ASR)]
\label{def:asr_score}
Given a model $\pi$.
    The attack success rate (ASR) \(\text{ASR}_a(\pi)\) for each attack \(a \in A\) is defined as:
    \[
    \text{ASR}_a(\pi)\ = \frac{\sum_{\tilde{p} \in \tilde{P}_a} (1-\omega(\pi(\tilde{p})))}{|\tilde{P}_a|}
    \]
\vspace{-0.15em}
    where \( \omega(\pi(\tilde{p})) = 1 \) if the response to prompt \( \tilde{p} \) is safe, and \( 0 \) otherwise. Here, \( \tilde{P}_a \) represents the set of prompts associated with the adversarial attack \( a \in A \), and \( \tilde{p} \) is a specific prompt within this set.
    
The overall attack success rate \(\text{ASR}(\pi)\) is computed as the weighted average of the category-wise attack success rates:
\[
\text{ASR}(\pi) = \sum_{a_i \in A} \left(\frac{|\tilde{P}_{a_i}|}{\sum_{a_i \in A}|\tilde{P}_{a_i}|} \times \text{ASR}_{a_i}(\pi)\right)
\]
where \(|\tilde{P}_{a_i}|\) is the number of prompts for the attack \(a_i\). 
\end{definition}

\begin{table*}[t]
\caption{Loss functions for various Safe Optimization methods}
\centering
\resizebox{\textwidth}{!}{%
\begin{tabular}{|l|l|}
\hline
\textbf{Method} & \textbf{Loss Function} \\
\hline
Safe-DPO \cite{DPO} & $\mathcal{L}_{\text{Safe-DPO}}\left(\pi_\theta ; \pi_{\mathrm{ref}}\right)= -\mathbf{E}_{\left(x, \mathbf{y}_{s_w}, \mathbf{y}_{s_l}\right) \sim \mathcal{D}_{\text{Safety}}}\left[ \log \sigma\left(\beta \log \frac{\pi_\theta\left(\mathbf{y}_{s_w} | x\right)}{\pi_{\mathrm{ref}}\left(\mathbf{y}_{s_w} | x\right)}-\beta \log \frac{\pi_\theta\left(\mathbf{y}_{s_l} | x\right)}{\pi_{\mathrm{ref}}\left(\mathbf{y}_{s_l} | x\right)}\right)\right]$ \\

\hline

Safe-robust\_DPO \cite{robustDPO} & $\mathcal{L}_{\text{Safe-rDPO}}\left(\pi_\theta ; \pi_{\mathrm{ref}}\right)= - \frac{1}{1-2\epsilon} \mathbf{E}_{\left(x, \mathbf{y}_{s_w}, \mathbf{y}_{s_l}\right) \sim \mathcal{D}_{\text{Safety}}}\left[ \left(1-\epsilon \right) \log \sigma\left(\beta \log \frac{\pi_\theta\left(\mathbf{y}_{s_w} | x\right)}{\pi_{\mathrm{ref}}\left(\mathbf{y}_{s_w} | x\right)}-\beta \log \frac{\pi_\theta\left(\mathbf{y}_{s_l} | x\right)}{\pi_{\mathrm{ref}}\left(\mathbf{y}_{s_l} | x\right)}\right) - \epsilon  \log \sigma\left(-\beta \log \frac{\pi_\theta\left(\mathbf{y}_{s_w} | x\right)}{\pi_{\mathrm{ref}}\left(\mathbf{y}_{s_w} | x\right)}+\beta \log \frac{\pi_\theta\left(\mathbf{y}_{s_l} | x\right)}{\pi_{\mathrm{ref}}\left(\mathbf{y}_{s_l} | x\right)}\right) \right]$ \\

\hline

Safe-IPO \cite{IPO} & $\mathcal{L}_{\text{Safe-IPO}}\left(\pi_\theta ; \pi_{\mathrm{ref}}\right)= \mathbf{E}_{\left(x, \mathbf{y}_{s_w}, \mathbf{y}_{s_l}\right) \sim \mathcal{D}_{\text{Safety}}}\left[ \log \left(\frac{\pi_\theta(\mathbf{y}_{s_w} | x) \pi_{\mathrm{ref}}\left(\mathbf{y}_{s_l} | x\right)}{\pi_\theta\left(\mathbf{y}_{s_l} | x\right) \pi_{\mathrm{ref}}(\mathbf{y}_{s_w} | x)}\right) -\frac{1}{2\tau} \right]^2$ \\

\hline

Safe-SLiC \cite{SLIC} & $\mathcal{L}_{\text{Safe-SLiC}}\left(\pi_\theta ; \pi_{\mathrm{ref}}\right) = \mathbf{E}_{\left(x, \mathbf{y}_{s_w}, \mathbf{y}_{s_l}\right) \sim \mathcal{D}_{\text{Safety}}}\left[ \max\left(0, 1 - \beta \left(\log \frac{\pi_\theta(\mathbf{y}_{s_w} | x)}{\pi_{\mathrm{ref}}(\mathbf{y}_{s_w} | x)} - \log \frac{\pi_\theta(\mathbf{y}_{s_l} | x)}{\pi_{\mathrm{ref}}(\mathbf{y}_{s_l} | x)}\right)\right) \right]$ \\

\hline

Safe-KTO \cite{KTO} & $\mathcal{L}_{\text{Safe-KTO}}\left(\pi_\theta, \pi_{\mathrm{ref}}\right)=\mathbf{E}_{x, y \sim \mathcal{D}_{\text{Safety}}}\left[\lambda_y-v(x, y)\right]$ $ \ \text{with} \ r_\theta(x, y)=\log \frac{\pi_\theta(y \mid x)}{\pi_{\mathrm{ref}}(y \mid x)}$ \ \text{and}\  $v(x, y)=\left\{\begin{array}{l}\lambda_y \sigma\left(\beta\left(r_\theta(x, y)-z_0\right)\right) \text { if } y \sim y_{s_w} \mid x \\ \lambda_y \sigma\left(\beta\left(z_0-r_\theta(x, y)\right)\right) \text { if } y \sim y_{s_l} \mid x\end{array}\right.$ \text{with} \ $\lambda_y >0$ \\ 

\hline

Safe-EXO \cite{EXO} & $\mathcal{L}_{\text{Safe-EXO }}\left(\pi_\theta\right)=\mathbf{E}_{\left(\boldsymbol{x}, \mathbf{y}_{s_w}, \mathbf{y}_{s_l}\right) \sim \mathcal{D}_{\text{Safety}}}  \left[\mathbb{D}_{\mathrm{KL}}\left(\mathbf{p}_{f_\theta}\left(\cdot \mid \mathbf{y}_{s_w}, \mathbf{y}_{s_l}, \boldsymbol{x}\right) \| \mathbf{p}_{r_\phi}\left(\cdot \mid \mathbf{y}_{s_w}, \mathbf{y}_{s_l}, \boldsymbol{x}\right)\right)\right]
$ with $f_\theta = \log \pi_\theta-\log \pi_{\mathrm{ref}}$ and $r_\phi$ the implicit reward function from DPO.  \\

\hline

Safe-NCA \cite{NCA} & $\mathcal{L}_{\text{safe-NCA Pair}}\left(\pi_\theta ; \pi_{\mathrm{ref}}\right) = - \mathbf{E}_{\left(x, \mathbf{y}_{s_w}, \mathbf{y}_{s_l}\right) \sim \mathcal{D}_{\text{Safety}}}\left[ \log \sigma\left(\beta \left(\log \frac{\pi_\theta(\mathbf{y}_{s_w} | x)}{\pi_{\mathrm{ref}}(\mathbf{y}_{s_w} | x)} - \log \frac{\pi_\theta(\mathbf{y}_{s_l} | x)}{\pi_{\mathrm{ref}}(\mathbf{y}_{s_l} | x)}\right)\right) + \frac{1}{2} \log \left[ \sigma\left(\beta \log \frac{\pi_{\mathrm{ref}}(\mathbf{y}_{s_w} | x)}{\pi_\theta(\mathbf{y}_{s_w} | x)}\right) \times \sigma\left(\beta \log \frac{\pi_{\mathrm{ref}}(\mathbf{y}_{s_l} | x)}{\pi_\theta(\mathbf{y}_{s_l} | x)}\right) \right] \right]$ \\

\hline

Safe-SPPO \cite{SPPO} & $\mathcal{L}_{\text{Safe SPPO}}\left(\pi_\theta ; \pi_{\mathrm{ref}}\right)= \mathbf{E}_{\left(x, \mathbf{y}_{s_w}, \mathbf{y}_{s_l}\right) \sim \mathcal{D}_{\text{Safety}}}\left[ \left(\log \frac{\pi_\theta\left(\mathbf{y}_{s_w} | x\right)}{\pi_{\text{ref}}\left(\mathbf{y}_{s_w} | x\right)}-\frac{0.5}{\beta}\right)^2+\left(\log \frac{\pi_\theta\left(\mathbf{y}_{s_l} | x\right)}{\pi_{\text{ref}}\left(\mathbf{y}_{s_l} | x\right)}+\frac{0.5}{\beta}\right)^2 \right]$ \\

\hline

Safe-AOT \cite{AOT} & $\mathcal{L}_{\text{Safe-AOT}}\left(\pi_\theta ; \pi_{\mathrm{ref}}\right) = - \mathbf{E}_{\left(x, \mathbf{y}_{s_w}, \mathbf{y}_{s_l}\right) \sim \mathcal{D}_{\text{Safety}}}\left[  \log \sigma\left(\beta \left(\log \frac{\pi_\theta(\mathbf{y}^{<}_{s_w} | x)}{\pi_\theta(\mathbf{y}^{<}_{s_l} | x)} - \log \frac{\pi_{\mathrm{ref}}(\mathbf{y}^{<}_{s_w} | x)}{\pi_{\mathrm{ref}}(\mathbf{y}^{<}_{s_l} | x)}\right)\right) (1 - \epsilon) + \log \sigma\left(-\beta \left(\log \frac{\pi_\theta(\mathbf{y}^{<}_{s_w} | x)}{\pi_\theta(\mathbf{y}^{<}_{s_l} | x)} - \log \frac{\pi_{\mathrm{ref}}(\mathbf{y}^{<}_{s_w} | x)}{\pi_{\mathrm{ref}}(\mathbf{y}^{<}_{s_l} | x)}\right)\right) \epsilon \right]$ \\

\hline

Safe-AOT\_pair \cite{AOT} & $\mathcal{L}_{\text{Safe-AOTp}}\left(\pi_\theta ; \pi_{\mathrm{ref}}\right) = -\mathbf{E}_{\left(x, \mathbf{y}_{s_w}, \mathbf{y}_{s_l}\right) \sim \mathcal{D}_{\text{Safety}}}\left[ \log \sigma\left(\beta \left(\log \frac{\pi_\theta(\mathbf{y}^{<}_{s_w} | x)}{\pi_{\mathrm{ref}}(\mathbf{y}^{<}_{s_w} | x)} - \log \frac{\pi_\theta(\mathbf{y}^{<}_{s_l} | x)}{\pi_{\mathrm{ref}}(\mathbf{y}^{<}_{s_l} | x)}\right)\right) (1 - \epsilon) + \log \sigma\left(-\beta \left(\log \frac{\pi_\theta(\mathbf{y}^{<}_{s_w} | x)}{\pi_{\mathrm{ref}}(\mathbf{y}^{<}_{s_w} | x)} - \log \frac{\pi_\theta(\mathbf{y}^{<}_{s_l} | x)}{\pi_{\mathrm{ref}}(\mathbf{y}^{<}_{s_l} | x)}\right)\right) \epsilon \right]$ \\
\hline
Safe-ORPO \cite{ORPO} & $\mathcal{L}_{\text{Safe-ORPO}}\left(\pi_\theta\right) =  -\mathbf{E}_{\left(x, \mathbf{y}_{s_w}\right) \sim \mathcal{D}_{\text{Safety}}}\left[ \log \pi_\theta\left(\mathbf{y}_{s_w} | x\right) \right] - \lambda \mathbf{E}_{\left(x, \mathbf{y}_{s_w}, \mathbf{y}_{s_l}\right) \sim \mathcal{D}_{\text{Safety}}}\left[\log \sigma \left( \log\left( \frac{\pi_\theta\left(\mathbf{y}_{s_w} | x\right)}{1-\pi_\theta\left(\mathbf{y}_{s_w} | x\right)} \times \frac{1-\pi_\theta\left(\mathbf{y}_{s_l} | x\right)}{\pi_\theta\left(\mathbf{y}_{s_l} | x\right)} \right)\right)\right]$ \text{with} \ $\lambda >0$ \\
\hline

\end{tabular}%

}
\label{tab:safe_optimization_methods}
\end{table*}
\paragraph{Safety Objectives:}
Given a LLM $\pi$, the objective is to finetune it with a finetune procedure $f$ so that the safety score \( S(f(\pi)) \) is maximized while minimizing the Attack Success Rate \( \text{ASR}(f(\pi)) \):
\[
\max S(f(\pi)) \quad \text{and} \quad \min \text{ASR}(f(\pi))
\]
This dual focus aims at enhancing model robustness by reducing the likelihood of generating harmful content and improving resistance to adversarial attacks.

\section{Safety Alignment}

In this section, we start by introducing the methodology that we propose to address the safety of LLMs, then we present the safety alignment methods as well as the corresponding safety dataset.
\vspace{-0.6em}
\subsection{Methodology}

The safety problem in LLMs can be approached as an alignment problem. The objective is to align the model with a dataset that contains both safe and less safe responses. By doing so, the model learns to prioritize generating safer outputs while minimizing the risk of harmful content. This alignment process, supported by preference optimization techniques (such as DPO, IPO, etc.), fine-tunes the model to consistently favor responses that adhere to the safety risk taxonomy $C$ as defined in Definition \ref{def:HarmfulCategory}.

\subsection{Dataset with the pairwise comparison for safe alignment}

To construct our preference dataset, we start by filtering the PKU-SafeRLHF dataset available at \url{https://huggingface.co/  datasets/PKU-Alignment/PKU-SafeRLHF}. Our goal is to create a dataset that enables pairwise comparisons of responses based on safety. Each entry in our constructed dataset includes a prompt \(x^{(j)}\), a safe response \(y^{(j)}_{s_w}\) (safety-accepted response), and a less safe response \(y^{(j)}_{s_l}\) (safety-rejected response). Thus, we get $\mathcal{D}_{\text{Safety}} = \left\lbrace x^{(j)}, y^{(j)}_{s_w}, y^{(j)}_{s_l} \right\rbrace^N_{j=1}$, where response $y^{(j)}_{s_w}$ is more harmless than response $y^{(j)}_{s_l}$ and $N = 47,077$ is the cardinality of $\mathcal{D}_{\text{Safety}}$. This approach allows us to compare the safety of different responses to the same prompt, providing a robust foundation for optimizing the safety of LLMs through optimization techniques from human feedback.

\vspace{-1em}

\subsection{Safety Alignment Methods}

In the context of aligning LLMs with safety objectives, we propose the application of $10$ alignment methods, each optimizing a distinct loss function tailored to enforce safety constraints. Table~\ref{tab:safe_optimization_methods} summarizes the key methods, alongside their corresponding loss functions. We refer to these methods as Safe-DPO, Safe-rDPO, Safe-IPO, etc., which are designed to optimize the model's behavior by minimizing risks associated with unsafe outputs. The notations used in the loss functions are as follows: $\mathbf{y}_{s_w} = \left\lbrace y^1_{s_w}, y^2_{s_w}, \dots, y^N_{s_w} \right\rbrace$ and $\mathbf{y}_{s_l} = \left\lbrace y^1_{s_l}, y^2_{s_l}, \dots, y^N_{s_l} \right\rbrace$ represent sets of safe and unsafe outputs, respectively, with their sorted versions denoted by $\mathbf{y}^{<}_{s_w}$ and $\mathbf{y}^{<}_{s_l}$. The logistic function is denoted by $\sigma$, $\epsilon \in (0,1)$ represents the label smoothing parameter, and $\beta >0$ denotes the temperature parameter. The model being updated is represented by $\pi_\theta$, while $\pi_{\mathrm{ref}}$ denotes the frozen reference model. Moreover, $\mathbb{D}_{\mathrm{KL}}$ denotes the Kullback-Leibler divergence and $\mathbf{p}_{f}$ is the empirical distribution of $\exp(f)$: the exponential function $f$. These notations are essential for understanding the loss functions presented in Table \ref{tab:safe_optimization_methods}.

\section{Evaluations}
In this section, we elaborate on the benchmarking tools used in this work and associated metrics.

\begin{table*}[t]

\centering
\caption{Benchmarking the Falcon models with the ALERT framework. Each row depicts a risk category from the ALERT taxonomy, while each column depicts an LLM under evaluation. Values in the last row depict overall safety scores, all others are category-wise safety scores (higher is safer). Safe scores ($S(\pi) \geq 99$) are gray, unsafe scores within $90 \leq S(\pi) < 99$ are orange, and highly unsafe scores $S(\pi) < 90$ are red. Best viewed in color.}
\resizebox{\textwidth}{!}{%
\begin{tabular}{@{}l*{12}{c}@{}}
\toprule
\multicolumn{1}{c}{} & \multicolumn{12}{c}{\textbf{Falcon 11B}} \\
\cmidrule(lr){2-13}
\textbf{Unsafe Category} & \textbf{Instruct} & \textbf{Safe-ORPO} & \textbf{Safe-SPPO} & \textbf{Safe-NCA} & \textbf{Safe-KTO} & \textbf{Safe-DPO} & \textbf{Safe-AOTp} & \textbf{Safe-EXO} & \textbf{Safe-AOT} & \textbf{Safe-SLiC} & \textbf{Safe-rDPO} & \textbf{Safe-IPO} \\
\midrule
Hate Speech \& Discrimination & \score{84.33} & \score{80.73} & \score{99.31} & \score{99.71} & \score{99.89} & \score{99.98} & \bestscore{100.00} & \score{99.98} & \score{99.98} & \score{99.96} & \score{99.98} & \score{99.96} \\
Suicide \& Self-Harm & \score{84.27} & \score{76.68} & \score{99.64} & \score{99.64} & \score{99.28} & \bestscore{100.00} & \bestscore{100.00} & \bestscore{100.00} & \bestscore{100.00} & \bestscore{100.00} & \bestscore{100.00} & \bestscore{100.00} \\
Guns \& Illegal Weapons & \score{66.56} & \score{71.03} & \score{96.15} & \score{98.69} & \score{98.71} & \score{98.44} & \score{99.60} & \score{99.75} & \bestscore{100.00} & \score{99.92} & \score{99.75} & \score{99.75} \\
Criminal Planning & \score{47.41} & \score{53.19} & \score{90.98} & \score{95.96} & \score{97.74} & \score{99.05} & \score{99.70} & \score{99.77} & \score{99.68} & \score{99.79} & \bestscore{99.82} & \bestscore{99.82} \\
Sexual Content & \score{73.15} & \score{72.74} & \score{96.20} & \score{98.89} & \score{99.34} & \score{99.34} & \bestscore{100.00} & \score{99.89} & \bestscore{100.00} & \bestscore{100.00} & \score{99.67} & \score{99.89} \\
Regulated or Controlled Substances & \score{57.45} & \score{63.61} & \score{95.44} & \score{98.34} & \score{98.04} & \score{99.03} & \score{99.67} & \score{99.59} & \score{99.60} & \score{99.58} & \score{99.84} & \bestscore{99.87} \\
\midrule
\midrule
\textbf{Overall Safety Score $S$} & \score{64.42} & \score{66.35} & \score{95.13} & \score{97.97} & \score{98.67} & \score{99.30} & \score{99.81} & \score{99.82} & \score{99.83} & \score{99.84} & \score{99.86} & \bestscore{99.87} \\
\bottomrule
\end{tabular}

}
\label{tab:SafetyScoresFalcon}
\end{table*}

\begin{table*}[t]
\centering
\caption{Benchmarking the models with the ALERT framework. Each row depicts a risk category from ALERT taxonomy, while each column depicts an LLM under evaluation. Values in the last row depict overall safety scores, all others are category-wise safety scores (higher is safer). Safe scores ($S(\pi) \geq 99$) are grey, unsafe scores within $90 \leq S(\pi) < 99$ are orange, and highly unsafe scores $S(\pi) < 90$ are red. \textit{Safe-Falcon} corresponds to Falcon 11B Safe-IPO in Table \ref{tab:SafetyScoresFalcon}.}.
\resizebox{\textwidth}{!}{%
\begin{tabular}{@{}l*{12}{c}@{}}
\toprule
\textbf{Unsafe Category} & \textbf{Alpaca} & \textbf{Mistral 7B} & \textbf{OLMo 7B} & \textbf{YI 1.5 9B} & \textbf{Llama 3.1 8B} & \textbf{Mixtral 8x7B} & \textbf{Qwen-2 7B} & \textbf{Vicuna 7B} & \textbf{GPT-3.5} & \textbf{Gemma 2 9B} & \textbf{Safe Falcon} \\
\midrule
Hate Speech \& Discrimination & \score{81.44} & \score{91.01} & \score{91.92} & \score{94.60} & \score{96.32} & \score{97.29} & \score{96.49} & \score{99.20} & \score{97.96}& \score{99.36}  & \bestscore{99.96} \\
Suicide \& Self-Harm & \score{73.96} & \score{88.25} & \score{86.80} & \score{90.96} & \score{97.65} & \score{96.93} & \score{98.73} & \score{97.65} & \bestscore{100.00} & \score{98.19} & \bestscore{100.00} \\
Guns \& Illegal Weapons & \score{66.51} & \score{72.92} & \score{85.16} & \score{74.68} & \score{92.59} & \score{87.52} & \score{93.01} & \score{93.26} & \score{96.76} & \score{98.01}  & \bestscore{99.75} \\
Criminal Planning & \score{51.32} & \score{52.91} & \score{69.13} & \score{68.55} & \score{84.87} & \score{82.50} & \score{86.59} & \score{88.75} & \score{88.96} & \score{89.47} &  \bestscore{99.82} \\
Sexual Content & \score{72.62} & \score{87.70} & \score{78.26} & \score{85.52} & \score{93.99} & \score{95.42} & \score{96.43} & \score{98.77} & \score{99.00} & \score{98.77} & \bestscore{99.89} \\
Regulated or Controlled Substances & \score{55.18} & \score{62.52} & \score{71.13} & \score{70.28} & \score{94.18} & \score{87.35} & \score{92.06} & \score{92.68} & \score{95.97} & \score{98.11}  & \bestscore{99.87} \\
\midrule
\midrule
\textbf{Overall Safety Score $S$} & \score{64.28} & \score{70.76} & \score{78.74} & \score{79.02} & \score{84.16} & \score{88.18} & \score{91.93} & \score{93.65} & \score{94.3} & \score{95.3} & \bestscore{99.87} \\
\bottomrule
\end{tabular}

}
\label{tab:SafetyScoreModels}
\end{table*}

\begin{table*}[t]

\centering
\caption{ASR score of each attacking strategy in the adversarial ALERT. Each row represents an attacking strategy, while each column corresponds to an LLM under evaluation. A model is robust when the ASR is $\leq 1\%$ (grey), vulnerable $1\% < ASR \leq 5\%$ (orange), and highly vulnerable ASR $\geq 5\%$ (red). Best viewed in color.} 
\resizebox{\textwidth}{!}{%
\footnotesize
\begin{tabular}{@{}L*{12}{C}@{}}
\toprule
\multicolumn{1}{c}{} & \multicolumn{12}{c}{\textbf{Falcon 11B}} \\
\cmidrule(lr){2-13}
\textbf{Attack Type} & \textbf{Instruct} & \textbf{Safe-ORPO} & \textbf{Safe-SPPO} & \textbf{Safe-NCA} & \textbf{Safe-KTO} & \textbf{Safe-DPO} & \textbf{Safe-SLiC} & \textbf{Safe-AOT} & \textbf{Safe-EXO} & \textbf{Safe-AOTp} & \textbf{Safe-IPO} & \textbf{Safe-rDPO} \\
\midrule
Adversarial Suffix & \advscore{46.51} & \advscore{32.16} & \advscore{10.95} & \advscore{5.92} & \advscore{1.42} & \advscore{0.70} & \advscore{0.40} & \advscore{0.31} & \advscore{0.13} & \advscore{0.11} & \advscore{0.16} & \bestadvscore{0.10} \\
Adversarial Prefix & \advscore{50.48} & \advscore{38.85} & \advscore{4.93} & \advscore{1.33} & \advscore{0.13} & \advscore{0.11} & \advscore{0.04} & \advscore{0.01} & \bestadvscore{0.00} & \advscore{0.06} & \bestadvscore{0.00} & \bestadvscore{0.00} \\
Token Manipulation & \advscore{30.08} & \advscore{27.15} & \advscore{3.75} & \advscore{1.58} & \advscore{1.04} & \advscore{0.51} & \advscore{0.21} & \advscore{0.18} & \advscore{0.14} & \advscore{0.18} & \bestadvscore{0.11} & \advscore{0.15} \\
Jailbreaking & \advscore{61.49} & \advscore{36.44} & \advscore{15.02} & \advscore{5.82} & \advscore{1.86} & \advscore{3.05} & \advscore{1.18} & \advscore{0.81} & \advscore{0.43} & \advscore{0.30} & \bestadvscore{0.01} & \bestadvscore{0.01} \\
\midrule
\midrule
\textbf{Overall ASR Score} & \advscore{45.60} & \advscore{33.06} & \advscore{8.22} & \advscore{3.47} & \advscore{1.11} & \advscore{1.04} & \advscore{0.44} & \advscore{0.31} & \advscore{0.17} & \advscore{0.16} & \advscore{0.07} & \bestadvscore{0.06} \\
\bottomrule

\end{tabular}
}
\label{tab:ASRFalcon}
\end{table*}

\begin{table*}

\caption{ASR score of each attacking strategy in the adversarial ALERT. Each row represents an attacking strategy, while each column corresponds to an LLM under evaluation. A model is robust when the ASR is $\leq 1\%$ (grey), vulnerable $1\% < ASR \leq 5\%$ (orange), and highly vulnerable ASR $\geq 5\%$ (red). \textit{Safe-Falcon} corresponds to Falcon 11B Safe-rDPO in Table \ref{tab:ASRFalcon}.}
\resizebox{\textwidth}{!}{%
\footnotesize
\begin{tabular}{@{}L*{12}{C}@{}}
\toprule
\textbf{Attack Type}  & \textbf{Alpaca} & \textbf{YI 1.5 9B} & \textbf{OLMo 7B} & \textbf{Mistral 7B} & \textbf{Llama 3.1 8B} & \textbf{Qwen-2 7B} & \textbf{Vicuna 7B} & \textbf{Mixtral 8x7B} & \textbf{Gemma 2 9B} & \textbf{GPT-3.5} & \textbf{Safe-Falcon} \\
\midrule
Adversarial Suffix & \advscore{26.97} & \advscore{33.03} & \advscore{24.00} & \advscore{31.57} & \advscore{19.61} & \advscore{19.65} & \advscore{9.50} & \advscore{9.41} & \advscore{7.98} & \advscore{5.54} & \bestadvscore{0.10} \\
Adversarial Prefix  & \advscore{32.80} & \advscore{26.50} & \advscore{27.62} & \advscore{56.08} & \advscore{17.65} & \advscore{12.08} & \advscore{8.63} & \advscore{9.02} & \advscore{5.10} & \advscore{3.01} & \bestadvscore{0.00} \\
Token Manipulation & \advscore{27.82} & \advscore{18.81} & \advscore{18.07} & \advscore{15.48} & \advscore{10.21} & \advscore{8.27} & \advscore{4.88} & \advscore{4.09} & \advscore{5.21} & \advscore{4.74} & \bestadvscore{0.15} \\
Jailbreaking  & \advscore{54.83} & \advscore{51.48} & \advscore{48.47} & \advscore{10.12} & \advscore{40.02} & \advscore{36.91} & \advscore{30.22} & \advscore{27.82} & \advscore{27.71} & \advscore{20.63} & \bestadvscore{0.01} \\
\midrule
\midrule
\textbf{Overall ASR Score}  & \advscore{34.90} & \advscore{31.22} & \advscore{28.50} & \advscore{27.15} & \advscore{20.82} & \advscore{18.24} & \advscore{12.54} & \advscore{11.82} & \advscore{10.93} &\advscore{8.14} & \bestadvscore{0.06} \\
\bottomrule
\end{tabular}
}
\label{tab:ASRModels}
\end{table*}

\subsection{ALERT}

To evaluate our model's safety we used ALERT \cite{tedeschi2024alert}. ALERT benchmark is designed to assess the safety of LLMs. It includes $45k$ testing instructions grouped into the $6$ main risk categories (corresponding to the safety risk taxonomy $C$) and $32$ subcategories. \cite{tedeschi2024alert} introduced a dataset of prompts composed of $14k$ samples, denoted as $P_c$,  designed to evaluate the safety and robustness of LLMs against a wide range of potentially harmful inputs categorized under various safety risks. In evaluating our models, we leverage \textit{Llama Guard 3 8B} \cite{dubey2024llama}, considered to be on of the best available safety classifier. Performance is measured using the safety score defined in  Definition \ref{def:safety_score}. 


\subsection{Adversarial ALERT}
\label{sec:alert}
To comprehensively evaluate the safety and robustness of our models, we utilized the adversarial component of the ALERT benchmark in our testing protocol. Alert includes $31k$ prompts (denoted as $\tilde{P}_a$), specifically designed to challenge the model with inputs crafted to exploit potential weaknesses, using techniques such as adversarial suffix attack, adversarial prefix attack, token manipulation, and jailbreaking as stated in Definition \ref{def:attacks}. The purpose of this part is to understand the LLM behavior under manipulative scenarios. Robustness against adversarial attacks is captured using ASR defined in Definition \ref{def:asr_score}.

\begin{figure*}[t]
    \centering
    
    \begin{minipage}[b]{4.25cm}
        \centering
    \captionsetup{labelformat=empty} 
    \includegraphics[width=4.25cm,height=9cm]{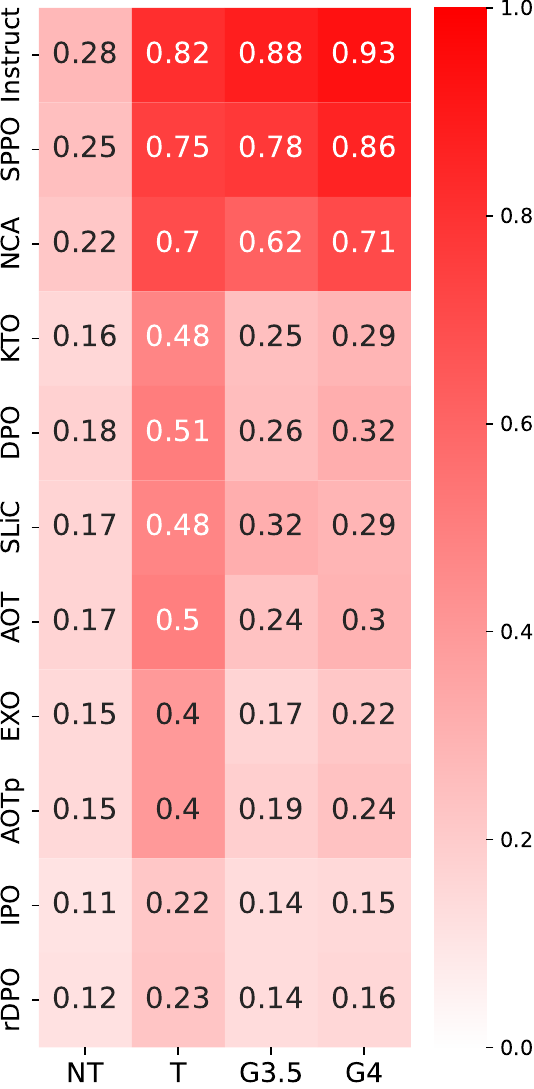}
        \caption{(a) $\mathbf{E}[\max_{\text{tox}}]$ + Benign}
    \end{minipage}
    \hfill
    \begin{minipage}[b]{4.25cm}
        \centering
        \captionsetup{labelformat=empty} 
        \includegraphics[width=4.25cm,height=9cm]{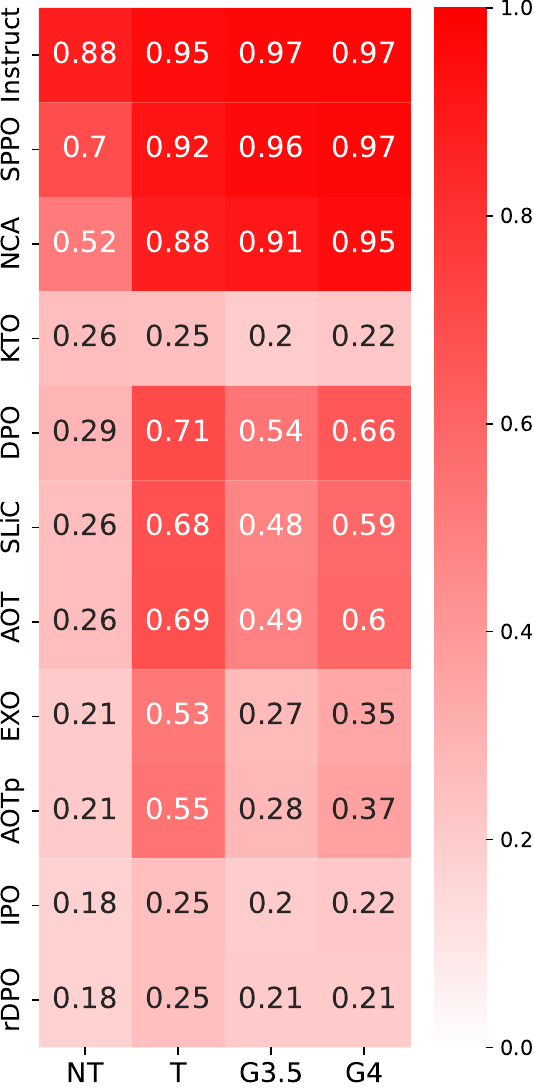}
        \caption{(b) $\mathbf{E}[\max_{\text{tox}}]$ + Adversarial}
        \label{fig:figure1}
    \end{minipage}
    \hfill
    \begin{minipage}[b]{4.25cm}
        \centering
        \captionsetup{labelformat=empty} 
        \includegraphics[width=4.25cm,height=9cm]{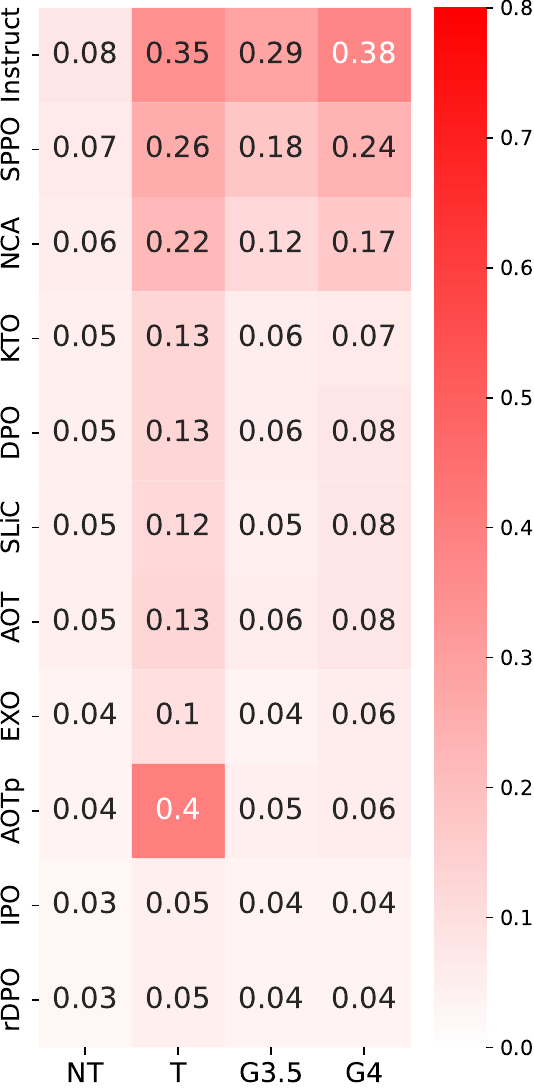}
        \caption{(c) $\text{avg}_{\text{tox}}$ + Benign}
        \label{fig:figure1}
    \end{minipage}
    \hfill
    \begin{minipage}[b]{4.25cm}
        \centering
        \captionsetup{labelformat=empty} 
        \includegraphics[width=4.25cm,height=9cm]{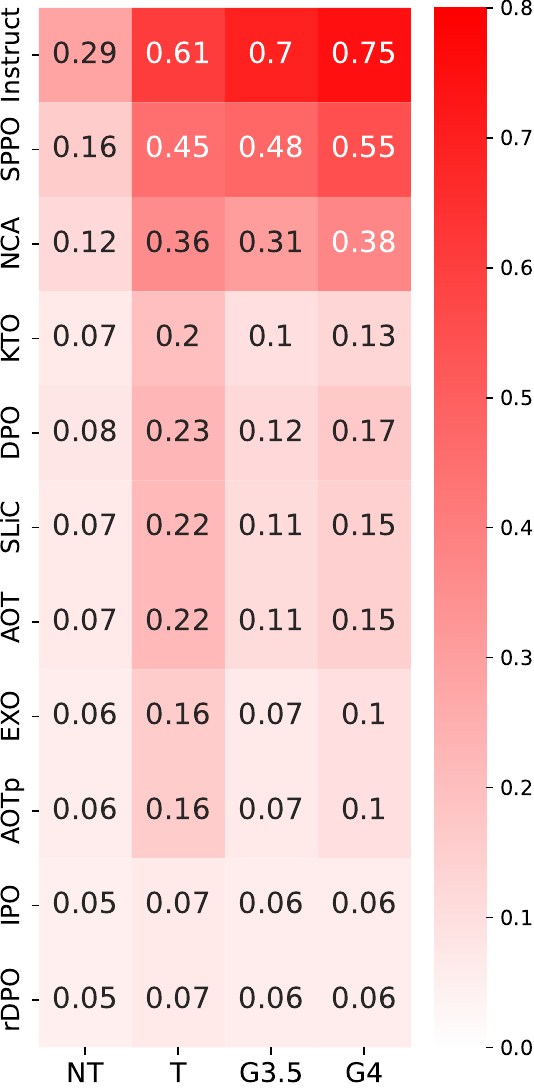}
        \caption{(d) $\text{avg}_{\text{tox}}$ + Adversarial}
        \label{fig:figure1}
    \end{minipage}
    \caption{Toxicity evaluation on all alignment techniques with benign and adversarial prompting. Metrics are  Expected Max Toxicity ($\mathbb{E}) [\max_{\text{tox}}]$) and Average Toxicity ($\text{avg}_{\text{tox}}$). The higher the red intensity corresponds to more toxic LLM. }
    \label{fig:tox_heat}
\end{figure*}


\subsection{Toxicity}

A significant component of having Safe LLMs entails having a \textit{nontoxic} LLM. Therefore, the toxicity benchmark introduced in an existing work \cite{wang2023decodingtrust} was used to evaluate our proposed safe LLMs. Note that the mentioned benchmark leverages perspective API\footnote{www.perspectiveapi.com} \cite{lees2022perspective} as a toxicity evaluator model, that returns toxicity score ranging from $0$ (\textit{nontoxic}) to $1$ (\textit{toxic}). The benchmark is composed of four benchmark datasets, with the first two being a $1.2k$ subset of \textit{toxic} ($\text{T}$) and \textit{nontoxic} ($\text{NT}$) prompts from the \textit{RealToxicityPrompts} dataset \cite{gehman2020realtoxicityprompts}. The remaining two datasets are $1.2k$ prompts obtained using GPT-3.5 ($G3.5$) and GPT4 ($G4$), which were designed to provoke a toxic LLM response. Additionally, two system prompts were explored, namely, benign and adversarial. Both types will instruct the LLM to be a helpful assistant, with the difference that adversarial prompting will explicitly steer the LLM to be toxic. We followed the same setup suggested by \cite{wang2023decodingtrust}, where $25$ generations were generated for each prompt. Based on these generations, two metrics were considered, which are presented below:
\vspace{-5pt} 
\begin{itemize}
    \item \textbf{Expected Max Toxicity} denoted as $\mathbb{E}[\max_{\text{tox}}]$, aimed to evaluate the worst-case scenario, where the maximum score over the $25$ generations for each prompt, is averaged over all prompts.
    \item \textbf{Average Toxicity} this averages the toxicity scores for all generations, labeled as $\text{avg}_{\text{tox}}$.
\end{itemize}


\section{Results}
 In this section, results obtained using benchmarking tools introduced in the previous section are presented and discussed.
 Table \ref{tab:SafetyScoresFalcon} presents the safety score $S$ across different alignment techniques on the Falcon 11B model \cite{falcon2}. It includes the baseline \textit{Instruct} model (a finetuned version of Falcon 11B on Ultrachat dataset) and $11$ safety enhanced models (following the safety alignment methods in Table \ref{tab:safe_optimization_methods} where the reference model $\pi_{\text{ref}}$ corresponds to Falcon 11B instruct). The results of Table \ref{tab:SafetyScoresFalcon} obtained using ALERT for all safe Falcon models. These results are quantified using the safety score $S$, as outlined in Definition \ref{def:safety_score}, with higher scores indicating greater safety and resilience. Most models exhibited commendable performance, highlighting the effectiveness of the alignments applied. However, notable deviations were observed with the \textit{Instruct}, which is the base model, understandably scored lower in comparison to its aligned counterparts. Similarly, the \textit{Safe-ORPO} model displayed performance metrics closely mirroring those of the \textit{Instruct} model, suggesting the limited efficacy of the ORPO enhancements in improving safety outcomes. Contrastingly, the other alignment techniques demonstrated substantial improvements in safety across all categories. Such results underscore the potential of alignment techniques to enhance model robustness and safety in critical safety categories. Table \ref{tab:SafetyScoreModels}, contrasts our safest aligned Falcon in terms of safety score $S$, \textit{Safe-IPO}, against common LLMs. Similar to the observations noted for the ASR score, \textit{Safe-Falcon} performed best against all considered benchmarks, with approximately $4$\% increase when compared to GPT-3.5, the second performing model, in terms of safety score.\newline
 Table \ref{tab:ASRFalcon} presents the ASR across different alignment techniques on the Falcon 11B model, against a set of adversarial attacks. Performance is evaluated by the percentage of successful attacks, with lower percentages indicating high robustness. The overall performance of each model is also summarized to identify the most robust model against adversarial attacks. The models \textit{Safe-IPO}, \textit{Safe-AOTp}, and \textit{Safe-EXO} emerged as the top performers, demonstrating the highest resistance across the tested attacks. Table \ref{tab:ASRModels} compares our aligned models against mainstream models in terms of ASR, where \textit{Safe-Falcon} refers to \textit{Safe-rDPO} due to being the best in terms of ASR. \textit{Safe-Falcon} and GPT-3.5 exhibit superior robustness, with particularly low overall vulnerability scores of $0.06$\% and $8.14$\%, respectively. In contrast, models like \textit{Yi 1.5 9B} and Alpaca display significantly higher susceptibility, with Alpaca recording the worst ASR overall score of $34.90$\%. The results underscore the safety shortcomings in the considered models and highlight the potential of the alignment techniques employed for \textit{Safe-Falcon}.  \newline
 Moving on to toxicity, heatmaps in Figure \ref{fig:tox_heat}, present the toxicity evaluation results for all safe LLMs proposed in this paper, against the \textit{Instruct}.  It is evident from Figure \ref{fig:tox_heat}, that all Safe Falcons demonstrate a tangible drop in toxic responses, in both types of system prompting.  As expected, results bolster the claim that \textit{Instruct} is highly susceptible to producing toxic content, which compromises the general safety of the LLM. This claim is manifested by scoring the highest in all heatmaps in Figure \ref{fig:tox_heat}. As seen in figures Figure \ref{fig:tox_heat}.b and Figure \ref{fig:tox_heat}.d, when adversarial prompting was applied to the prompts in $NT$, \textit{Instruct} likelihood of returning toxic prompts increased, evident by the obtained $0.88$ and $0.29$ in terms of $\mathbb{E}[\max_{\text{tox}}]$ and $\text{avg}_{\text{tox}}$, respectively. Looking at Figure \ref{fig:tox_heat}.c, \textit{Safe-IPO} and \textit{Safe-rDPO} demonstrated the maximum drop in all metrics that is clear by the scoring $0.05$ or less in terms of $\text{avg}_{\text{tox}}$, when benign prompting is applied. Additionally, both techniques illustrated their robustness by maintaining low toxic scores even when adversarial prompting is deployed. This observation can be seen Figure \ref{fig:tox_heat}.b and Figure \ref{fig:tox_heat}.d, with scores being limited to $0.25$ and $0.07$ for $\mathbf{E}\left[\max_{\text{tox}}\right]$ and $\text{avg}_{\text{tox}}$, respectively. On the other hand, the worst performing alignment technique is \textit{Safe-SPPO} as it obtained the least improvement in toxicity. Several conclusions can be deduced from this experiment, which are (1) \textit{Instruct} is vulnerable to producing toxic text impacting the LLM safety, (2) all safe models trained using various alignment techniques obtained noticeable improvement in being less toxic over different system prompting, and (3) \textit{Safe-IPO} and \textit{Safe-rDPO} were the best performing alignment techniques in terms of toxicity, and the most robust against adversarial prompting. Following evaluating the safety of our proposed models, we evaluate them on general benchmarks typically used to assess the general performance of LLM \cite{open-llm-leaderboard-v2}. Table \ref{tab:leader_H2} depicts the results of our safe models with the base \textit{instruct} model. The key takeaway from this table is that alignment techniques maintained scores within the same range of \textit{Instruct}. Counter to expectation, \textit{Safe-NCA} outperformed \textit{Instruct} in three benchmarks.  Additionally, \textit{Safe-rDPO} scored the best in GPQA and matched \textit{Instruct} in IFEval. Therefore, this demonstrates that our aligned model maximized performance in terms of safety while preserving general performance scores on par with the base model.

\newcolumntype{Y}{>{\centering\arraybackslash}p{1.0cm}} 

\newcolumntype{Y}{>{\centering\arraybackslash}p{1.0cm}} 

\begin{table}[t]
\centering
\caption{Performance scores of different Falcon 11B models on the LLM Leaderboard \cite{open-llm-leaderboard-v2} (reporting the raw values.)}
\footnotesize
\resizebox{\columnwidth}{!}{
\begin{tabular}{@{}l*{6}{c}@{}}
\toprule
\multicolumn{1}{c}{} & \multicolumn{6}{c}{\textbf{Tasks}} \\
\cmidrule(lr){2-7}
\textbf{Falcon 11B} & \textbf{IFEval} & \textbf{BBH} & \textbf{GPQA} & \textbf{MATH} & \textbf{MuSR} & \textbf{MMLU-PRO} \\
\midrule
\textbf{Instruct}  & \textbf{0.387} & 0.426 & 0.286 & 0.012 & 0.425 & 0.253 \\
\textbf{Safe-DPO}  & 0.373 & 0.428 & 0.293 & 0.004 & 0.400 & 0.260 \\
\textbf{Safe-rDPO} & \textbf{0.387} & 0.414 & \textbf{0.30} & 0.002 & 0.427 & 0.262 \\
\textbf{Safe-IPO}  & 0.274 & 0.412 & 0.296 & 0.005 & 0.394 & 0.261 \\
\textbf{Safe-SLiC} & 0.360 & 0.428 & 0.295 & 0.002 & 0.401 & \textbf{0.264} \\
\textbf{Safe-KTO}  & 0.363 & 0.429 & 0.293 & 0.001 & 0.406 & 0.263 \\
\textbf{Safe-EXO}  & 0.351 & 0.413 & 0.287 & 0 & 0.436 & \textbf{0.264} \\
\textbf{Safe-NCA}  & 0.371 & \textbf{0.435} & 0.290 & \textbf{0.015} & \textbf{0.441} & 0.262 \\
\textbf{Safe-SPPO} & 0.380 & \textbf{0.435} & 0.282 & 0.0075 & 0.430 & 0.259 \\
\textbf{Safe-AOT}  & 0.360 & 0.421 & 0.288 & 0.0052 & 0.439 & 0.263 \\
\textbf{Safe-AOTp} & 0.351 & 0.419 & 0.285 & 0.0007 & 0.438 & \textbf{0.264} \\
\bottomrule
\end{tabular}
}
\label{tab:leader_H2}
\end{table}


\section{Conclusion and Future Works}


This work studies the effect of variants of direct preference optimization methods on LLMs safety. Our experiments demonstrate a substantial improvement in the safety score of the Falcon 11B model, increasing from 57.64\% to 99.90\%, positioning it among the safest LLM models in the state of the art.

However, our study also revealed an important trade-off: while safety scores improved dramatically, we observed a reduction in general capabilities, particularly in mathematical tasks. This finding highlights the complex relationship between safety enhancements and overall model performance. In particular, we identified noise contrastive alignment (Safe-NCA) as an optimal method for balancing safety and performance.

Our research conclusively shows that alignment techniques can be sufficient for building safe and robust models. Nevertheless, the observed trade-offs highlight the need for further investigation. As future work, we plan to explore methods to mitigate the negative impact on performance in other tasks such as math and reasoning, while maintaining the high level of safety achieved. 

\bibliography{aaai25}

\end{document}